%% file: main.tex
\documentclass[10pt,twocolumn,letterpaper]{article}

\usepackage[pagenumbers]{cvpr} 


\input{lib}

\definecolor{cvprblue}{rgb}{0.21,0.49,0.74}
\usepackage[pagebackref,breaklinks,colorlinks,allcolors=cvprblue]{hyperref}

\def\paperID{1089} 
\def\confName{CVPR}
\def\confYear{2025}

\title{Recovering Dynamic 3D Sketches from Videos}

\author{
Jaeah Lee \hspace{0.05\linewidth} Changwoon Choi \hspace{0.05\linewidth} Young Min Kim \hspace{0.05\linewidth} Jaesik Park\thanks{Corresponding author.}\\[2pt]
{Seoul National University, Republic of Korea}\\
{\tt\small hayanz@snu.ac.kr \hspace{0.01\linewidth} changwoon.choi00@gmail.com  \hspace{0.01\linewidth} \{youngmin.kim, jaesik.park\}@snu.ac.kr}
}

\begin{document}

\maketitle
\input{sec/0_abstract}    
\input{sec/1_intro}
\input{sec/2_related_work}
\input{sec/3_method}
\input{sec/4_experiment}
\input{sec/5_conclusion}
\input{sec/9_ack}
{
    \small
    \bibliographystyle{ieeenat_fullname}
    \bibliography{refs}
}
\input{sec/X_suppl}

\end{document}

%% file: lib.tex
\usepackage[accsupp]{axessibility}
\usepackage{graphicx}
\usepackage{multirow}
\usepackage{tabularx}
\usepackage[ruled, vlined]{algorithm2e} 
\usepackage{url}                        
\usepackage{booktabs}                   
\usepackage{amsfonts}                   
\usepackage{nicefrac}                   
\usepackage{microtype}                  
\usepackage{xcolor}                     
\usepackage{mathtools}
\usepackage{amsmath}
\usepackage{amssymb}
\usepackage{xspace}
\usepackage{pifont}
\usepackage{cuted}
\usepackage{etoolbox}
\usepackage{makecell}
\usepackage{natbib}

\newcommand{\Sec}{Sec.}
\newcommand{\fig}{Fig.}
\newcommand{\tab}{Tab.}
\newcommand{\Eq}{Eq.}
\newcommand{\App}{Appendix}



%% file: sec/0_abstract.tex
\begin{abstract}

Understanding 3D motion from videos presents inherent challenges due to the diverse types of movement, ranging from rigid and deformable objects to articulated structures.
To overcome this, we propose Liv3Stroke, a novel approach for abstracting objects in motion with deformable 3D strokes.
The detailed movements of an object may be represented by unstructured motion vectors or a set of motion primitives using a pre-defined articulation from a template model.
Just as a free-hand sketch can intuitively visualize scenes or intentions with a sparse set of lines, we utilize a set of parametric 3D curves to capture a set of spatially smooth motion elements for general objects with unknown structures.
We first extract noisy, 3D point cloud motion guidance from video frames using semantic features, and our approach deforms a set of curves to abstract essential motion features as a set of explicit 3D representations.
Such abstraction enables an understanding of prominent components of motions while maintaining robustness to environmental factors.
Our approach allows direct analysis of 3D object movements from video, tackling the uncertainty that typically occurs when translating real-world motion into recorded footage.
The project page is accessible via: \url{https://jaeah.me/liv3stroke_web}.

\end{abstract}

%% file: sec/1_intro.tex
\section{Introduction}


\begin{figure}
    \centering
    \includegraphics[trim={0mm 2mm 0mm 5mm}, clip, width=0.97\linewidth]{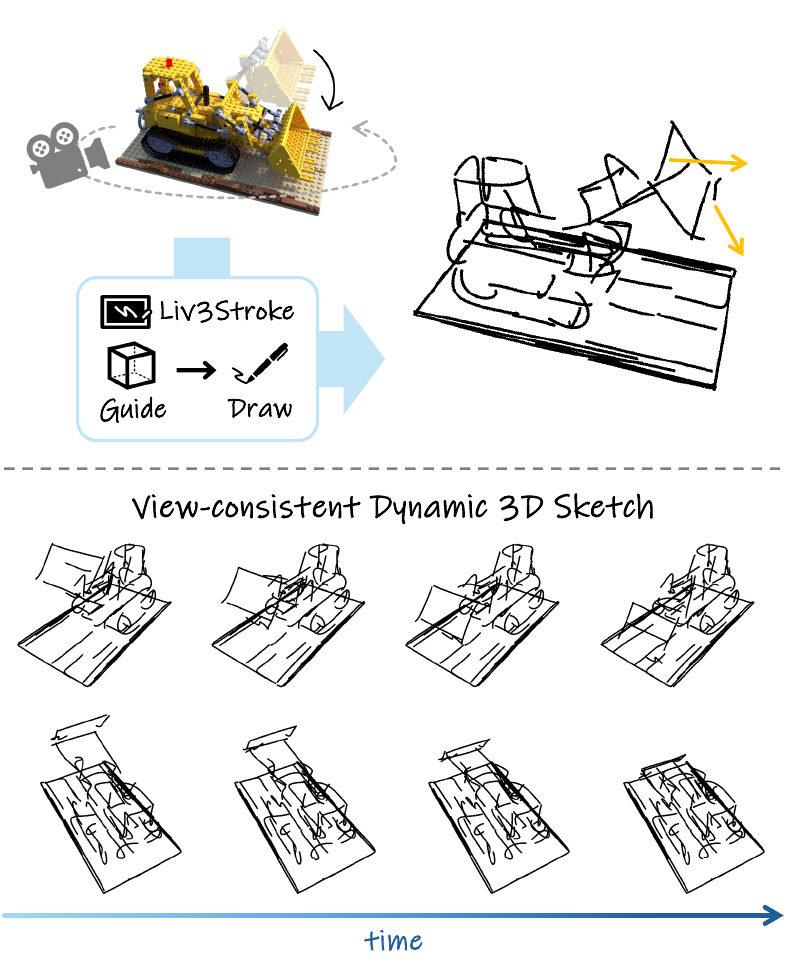}
    \vspace{-7pt}
    \caption{
    Liv3Stroke is a novel approach that compactly represents object movements using deformable 3D strokes from videos.
    Our method achieves view-consistent dynamic sketch reconstruction by shifting and deforming the shape of each stroke.
    }
    \label{fig:teaser}
    \vspace{-10pt}
\end{figure}

Tracking 3D movement in video is subject to inherent ambiguity.
The 3D motions usually exhibit various forms, including unidentified rigid, deformable, and articulated objects.
A video footage observes the 3D motions that are projected onto 2D frames of a moving camera, which further complicates the process of extracting motion components.
There have been attempts to understand locally smooth motion patterns, such as motion factorization~\cite{noguchi2022watch} or segmentation~\cite{yang2023movingparts}.
Recent approaches on novel-view synthesis extend the radiance field formulation into a more general form of dynamic scenes with dense motion fields~\cite{pumarola2021d, wu20244d, yang2023deformable3dgs, li2023dynibar}.
However, this formulation solves for complex high-dimensional variables given an under-constrained set-up, and often produces erroneous results that are further deteriorated when the scene is subject to appearance variations.

We find the key insights to express 3D motion through \textit{abstraction} in the form of sketches.
Sketches serve as effective tools for compactly visualizing scenes or ideas~\cite{gryaditskaya2019opensketch, hertzmann2020line}.
While they are subjective expressions, often created by an artist, recent works~\cite{vinker2022clipasso, vinker2023clipascene} have shown that we can generate sketches directly from images guided by visual features.
They define a perceptual loss in a latent space incorporating deep neural networks and successfully generate abstract sketches.
Interestingly, the perceptual loss alleviates enforcing pixel-wise matches in image space, and it robustly extracts meaningful features even under ambiguous situations without precise alignment.
Inspired by this, subsequent works incorporate similar formulations to retrieve view-consistent structural information in 3D~\cite{choi20243doodle} or compact representation for dynamic videos~\cite{zheng2024sketch, gal2024breathing}.
Likewise, we propose that deformable sketch lines can be a flexible yet compact parameterization representing the arbitrary topology of locally smooth 3D motions in videos.

Meanwhile, 3D strokes have the potential to describe the dense field. We can express a 3D concept in a sparse and abstracted form as a wire-like structure~\cite{Tojo2024Wireart, qu2023dreamwire}.
Starting from this property, there have been attempts to apply 3D curves in sketch-based modeling~\cite{luo2021simpmodeling} or surface editing~\cite{yu2022piecewise} to depict intended details.
Different from these works, we focus on the ``deformability" of each stroke, which enables effectively capturing 3D key features of diverse movements.

To align sketches with dynamic scenes, we define a sketch as a set of deformable 3D strokes and utilize vectorized curves (cubic B\'ezier curves) for the expression.
Using the editing capabilities of vector graphics, we can effectively convey movements by shifting stroke positions and their control points.
%
Before reconstructing moving sketches, we compute a dense guiding motion field since it is challenging to directly align strokes with input video frames.
We first reconstruct 3D motion guidance, which is a dynamic point cloud with a deformation network.
We propose to optimize motion guidance with rendering loss in perceptual space.
This guidance serves as a rough initial 3D motion and location for dynamic 3D strokes.
%
Based on this approach, we fit 3D strokes into movements.
We represent motions by individually relocating each curve and adjusting its control points.
We minimize the gap between stroke deformation and motion through a coarse-to-fine approach, which enables us to maintain the core shape throughout the movement.

We show its performance in intuitively expressing smooth motion from video frames.
Our approach enables to draw core view-consistent object motions in 3D space, and can capture the object's overall shape throughout the entire video sequence while being robust to external factors.
We present that our approach can render diverse movements, even if capturing 3D motion from videos is challenging due to the difference between the velocity of camera movements and motions~\cite{gao2022monocular}.


To summarize, we introduce the following contributions:

\begin{itemize}
    \item{We propose Liv3Stroke, which is the first approach to reconstruct dynamic sketches in 3D space and abstract movements with sparse strokes.}
    \item{We define each stroke of the sketch by B\'ezier curves, and motions are represented by changing each curve’s location or shape.}
    \item{Our approach can concisely express core structures and motions from natural videos, including monocular video with moving camera poses.}
    \item{Our approach draws view-consistent object motions in 3D space and captures their key features while being less affected by environmental factors.}
\end{itemize}

%% file: sec/2_related_work.tex
\section{Related Work}

\paragraph{Dynamic 3D Scene Reconstruction}
The emergence of neural radiance fields~\cite{mildenhall2021nerf} has catalyzed significant advancements in photorealistic 3D reconstruction from multi-view images.
Spontaneously, there have been attempts for dynamic scene reconstruction in 3D space.
Fridovich \etal~\cite{fridovich2023k} and Cao and Johnson~\cite{cao2023hexplane} represent dynamic 3D scenes by Eulerian motion field defined in a space-time 4D grid.
Also, there are some works~\cite{kocabas2024hugs, jiang2022neuman, hu2024gauhuman, xu2024gaussian, hu2024gaussianavatar, qian2024gaussianavatars} using template model that is specialized in specific objects, such as a parametric human model.

Meanwhile, most existing works~\cite{pumarola2021d, Li_2021_CVPR, Liu_2023_CVPR, park2021nerfies, hypernerf} follow a common pattern; they reconstruct dynamic 3D scenes by optimizing canonical 3D scenes and warp them by learnable deformation fields.
Our method follows a similar approach yet focuses on concisely representing motions in 3D space.
We aim to recover dynamic sketches by learning the deformation of strokes at each timestep.
\vspace{-10pt}

\paragraph{Stroke-Based Representation}
Stroke-based representation expresses scenes with a few strokes.
It is one of the simplest sketch representations yet effective in conveying the essential structures and semantics of target objects.
There are learning-based methods to synthesize sketches from images~\cite{muhammad2018learning} by training neural networks on limited image-sketch paired datasets.
Optimization-based methods~\cite{vinker2022clipasso, vinker2023clipascene} utilize strong prior vision language model~\cite{radford2021learning}, and they enable the generation of sketches without training on specific classes.
Recently, some works have tackled extending stroke-based representation to the video domain.
Gal~\etal~\cite{gal2024breathing} try to animate sketches given a text prompt, and Zheng \etal~\cite{zheng2024sketch} reconstruct abstract dynamic 2D sketches from videos.
Furthermore, 3Doodle~\cite{choi20243doodle} and EMAP~\cite{li20243d} reconstruct 3D strokes from multi-view images, enabling better representation of 3D shapes compared to 2D strokes.
However, these works are still limited since they mainly target stationary scenes.

In this paper, we aim to reconstruct motions as sketches in 3D space.
We capture dynamic motions from videos by positioning and deforming view-consistent strokes.

%% file: sec/3_method.tex
\begin{figure*}[ht]
    \centering
    \includegraphics[trim={0mm 13mm 0mm 13mm}, clip, width=1.0\textwidth]{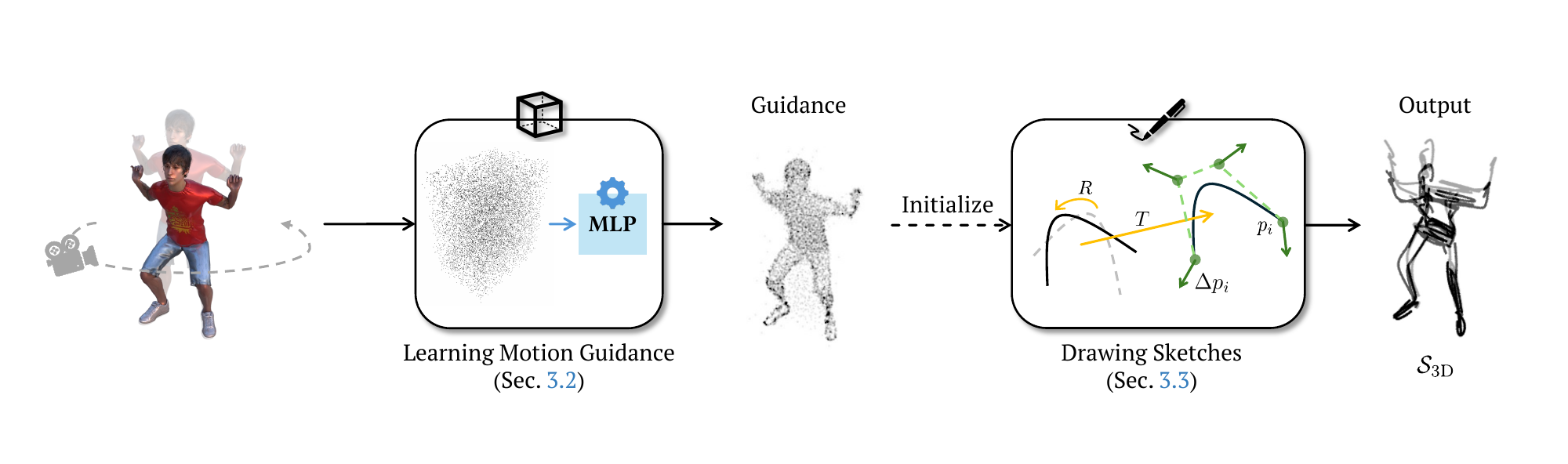}
    \caption{Method overview.
    We first learn 3D motion guidance from video frames, which are defined as a set of point cloud.
    Based on this, we can initialize approximated stroke position and motions.
    We represent movement by transforming each individual stroke through rotation $R$ and translation $T$, and adjusting its control points $\{p_i\}$ with displacement $\{\Delta p_i\}$, thereby reconstructing a dynamic 3D sketch $\mathcal{S}_{\mathrm{3D}}$.
    }
    \label{fig:method_overview}
\end{figure*}
\vspace{-5pt}

\section{Method}

In this section, we explain how to convey motions with 3D strokes. Our overall method is shown in \fig~\ref{fig:method_overview}.
We introduce our compact sketch representation for motion abstraction in \Sec~\ref{subsec:representation}, and \Sec~\ref{subsec:guiding} details the process of guiding rough motions in 3D space.
Based on the motion guidance, we describe dynamic sketches through the process described in \Sec~\ref{subsec:drawing}.

\input{sec/3-1_representation}
\input{sec/3-2_motion_extraction}
\input{sec/3-3_drawing}

%% file: sec/3-1_representation.tex
\subsection{Sketch Representation}\label{subsec:representation}

We define a \textit{sketch} as $n$ black strokes on the white background.
Each stroke is represented as a 3D cubic B\'ezier curve, which is defined with four control points $s_i=\{p_i^j\}_{j=0}^3$, $p_i^j\in\mathbb{R}^3$.
To simplify the process, we use fully opaque sketch lines (\ie, fix the opacity as 1), and only adjust the position of control points.

When we project a 3D B\'ezier curve to the image plane, the represented spline is a 2D rational B\'ezier curve.
However, by assuming that the camera is located sufficiently far from the target object, resulting in negligible perspective distortion, the perspective projection can be approximated as orthographic projection following~\cite{choi20243doodle}.
Hence, we can consider the projected curve as a general 2D B\'ezier curve, and use an existing 2D differentiable rasterizer~\cite{Li:2020:DVG} $\mathcal{R}$ to render the projected curves.
Each sketch frame $\mathcal{S}$ is mathematically expressed as follows:
\begin{equation}
    \mathcal{S} = \mathcal{R}(\Psi(\mathcal{S}_{\mathrm{3D}}, M, K)),
\end{equation}
where $\Psi$ denotes the projection to a 2D image plane with the extrinsic matrix $M$ and the intrinsic matrix $K$, and $\mathcal{S}_{\mathrm{3D}}=\{s_{i}\}_{i=1}^{n}$ is the 3D sketch defined by a set of curves.
Then, we represent the interested object's movements by shifting each curve and its control points.

%% file: sec/3-2_motion_extraction.tex
\subsection{Learning 3D Motion Guidance}\label{subsec:guiding}

\begin{figure}
    \centering
    \includegraphics[trim={2mm 2mm 2mm 2mm}, clip, width=1.0\linewidth]{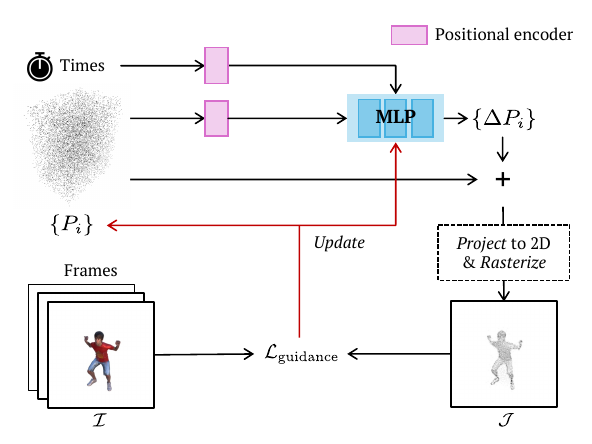}
    \caption{Framework for outlining 3D motion.
    To get the motion layout, we use a 3D point cloud and compute movements by repositioning its points.
    An MLP acts as the function that estimates motion $\{\Delta P_{i}\}$ across the provided video frames.
    }
    \label{fig:method_3dmotion}
    \vspace{-10pt}
\end{figure}

While existing dynamic scene reconstruction methods target dense photorealistic representation through direct optimization, our goal of compact abstraction demands understanding of the object's essential structure and motion patterns, which cannot be solved by a simple transfer from dense representation.
Hence, we first compute approximated 3D motion guidance using low-resolution images.
This then helps us determine how the strokes should be located, move, and flow.
The detailed framework of this stage is shown in \fig~\ref{fig:method_3dmotion}.

We represent the scene using a point cloud $\{P_{i}\}$ due to their efficiency and ease of manipulation.
The motion in the scene is then visualized through the movement of these point cloud elements.
Like the prior work~\cite{pumarola2021d}, we implement an MLP-based deformation network that takes time and position encodings as input to estimate spatial deformations at each time step.
We use the same positional encoder as in ~\cite{mildenhall2021nerf} where $\gamma(p) = (\sin(2^l\pi p), \cos(2^l\pi p))_{l=0}^{L-1}$.
The projected point is rasterized as a black Gaussian unit.
We describe the details of the rasterization process in \App.

Meanwhile, we cannot directly compare a projected image $\mathcal{J}$ with the corresponding RGB frame $\mathcal{I}$ since the rendered ouput $\mathcal{J}$ is an image intensity, as shown in \fig~\ref{fig:method_3dmotion}, not a natural image as $\mathcal{I}$.
Therefore, instead of using pixel-wise loss, we utilize LPIPS loss~\cite{zhang2018perceptual} to get perceptual alignment quality and assess the structural difference:
\begin{equation}\label{eq:guiding_frame}
    \mathcal{L}_{\mathrm{frame}}^{\mathrm{g}}=\rho(\mathrm{LPIPS}(\mathcal{I}, \mathcal{J}), \alpha, c),
\end{equation}
where $\mathcal{I}$ and $\mathcal{J}$ are the training image and a rasterized point cloud. $\rho(\mathbf{x}, \alpha, c)$ is a robust function~\cite{barron2019general} that stabilizes the optimization process by reducing the impact of outliers.
The parameters $\alpha$ and $c$ are set to 1 and 0.1, respectively.

\begin{figure}
    \centering
    \includegraphics[trim={0mm 0mm 0mm 0mm}, clip, width=0.85\linewidth]{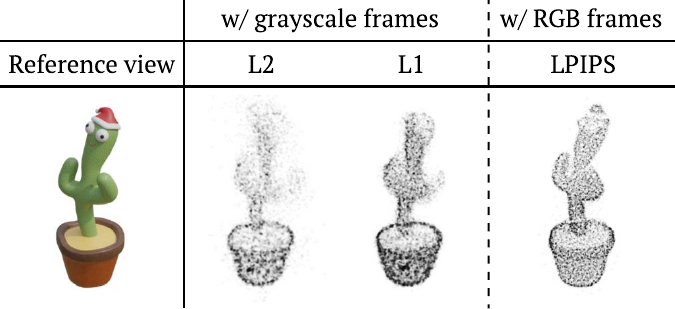}
    \caption{Comparison of loss functions.
    Compared to pixel-wise losses (L2 and L1 function), LPIPS loss provides stricter structural guidance when computing differences between two images, even with RGB frames.
    }
    \label{fig:method_guiding_loss}
    \vspace{-10pt}
\end{figure}

As shown in \fig~\ref{fig:method_guiding_loss}, LPIPS loss preserves the structural integrity of the cactus better than L1 or L2 loss.
While they produce noisy and scattered representations even with grayscale frames, LPIPS maintains clearer object boundaries and more coherent shape details directly from images.

To achieve smooth motion, we introduce two regularization terms in \Sec~\ref{subsec:guiding} and \Sec~\ref{subsec:drawing}: a velocity continuity term that ensures stable transitions and a shape stability term that prevents excessive deformation.
The former is defined as:
\begin{equation}\label{eq:guiding_temporal}
\mathcal{L}_{\mathrm{temp}}^{\mathrm{g}}=\lVert\frac{\Delta P_{\mathrm{3D}}^{t} - \Delta P_{\mathrm{3D}}^{t'}}{t-t'}\rVert_{2},
\end{equation}
where $t'$ is a neighboring time step of $t$, randomly sampled within the range $[t-dt, t]$ with $dt$ representing the temporal interval between adjacent frames, and $\{\Delta P_{\mathrm{3D}}^{t}\}$ is changes of point locations at time step $t$.

For the latter, we need stronger constraints in this step compared to drawing sketches (\Sec~\ref{subsec:drawing}) since we aim to get a ``dense'' motion field.
Hence, we use L1 loss to ensure a more stable shape throughout the entire sequence.
We compute the rigid transformation between time step $t$ and $t'$, then compute the following regularization term:
\begin{equation}\label{eq:guiding_rigid}
    \mathcal{L}_{\mathrm{rigid}}=\|\smash{^{q}R_{t\rightarrow t'}}-\smash{^{q}I}\|_{1} + \|T_{t\rightarrow t'}\|_{1},
\end{equation}
where $R_{t\rightarrow t'} \in \mathrm{SO}(3)$ and $T_{t\rightarrow t'} \in \mathbb{R}^{3}$ denote the rigid rotation and translation, respectively.
We use the well-known algorithm described in \cite{besl1992method} and \cite{horn1987closed} to compute these terms.
The expression $\smash{^{q}}M$ is a corresponding quaternion of the matrix $M\in\mathrm{SO}(3)$ to avoid gimbal lock~\cite{shoemake1985animating}.
The overall objective function is as follows:
\begin{equation}\label{eq:guiding_total}
 \mathcal{L}_{\mathrm{guidance}}=\omega_{\mathrm{f}}\mathcal{L}_{\mathrm{frame}}^{\mathrm{g}}+\omega_{\mathrm{t}}\mathcal{L}_{\mathrm{temp}}^{\mathrm{g}}+\omega_{\mathrm{r}}\mathcal{L}_{\mathrm{rigid}}.
\end{equation}
Following this approach, we can get the hint of 3D movements and initialize canonical stroke locations, \ie, positions before deformation.
We further discuss the effect of this obtained guidance in \Sec~\ref{subsec:exp_abl}.

%% file: sec/3-3_drawing.tex
\subsection{Drawing Dynamic Sketches}\label{subsec:drawing}

\begin{figure*}
    \centering
    \includegraphics[width=1.0\textwidth]{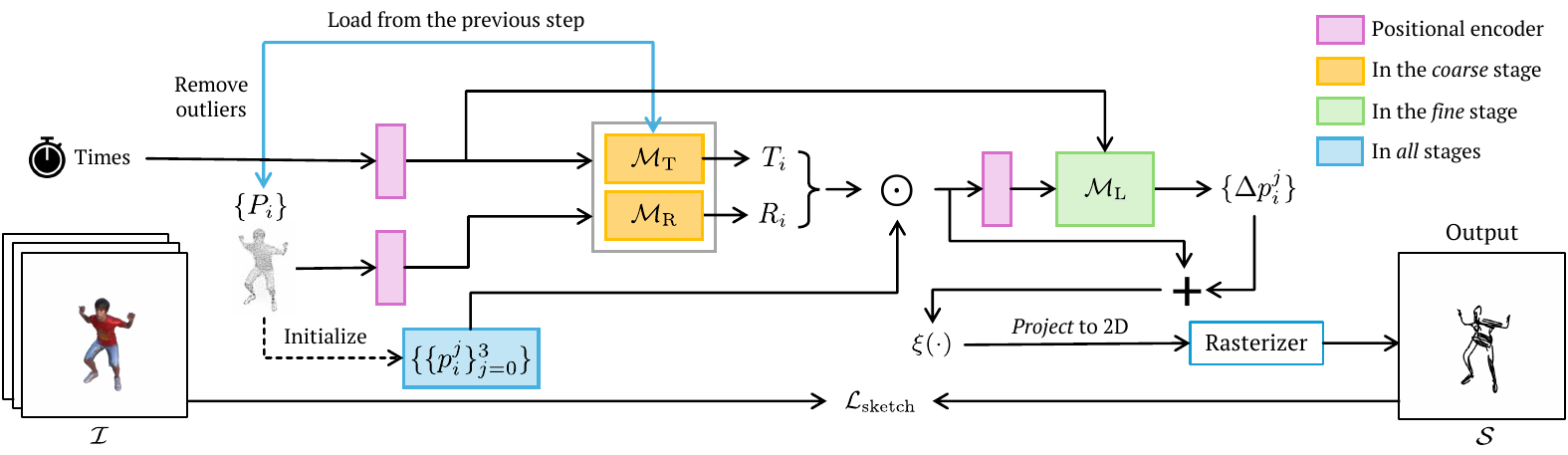}
    \caption{Pipeline for rendering sketches.
    We extract the motion as sketches by learning the deformation of each stroke.
    To achieve this, we separate stroke motions as (1) per-stroke transformations, which are composed of rotation $R_{i}$ and translation $T_{i}$ and (2) shifting each control point of the stroke $\{\Delta p_{i}^{j}\}$.
    We use MLPs, $\mathcal{M}_{\mathrm{R}}$, $\mathcal{M}_{\mathrm{T}}$, and $\mathcal{M}_{\mathrm{L}}$, as the function of deformation at the given time step.
    }
    \vspace{-5pt}
    \label{fig:method_sketch}
\end{figure*}

Based on the extracted motion guidance, we draw motions with 3D strokes by learning the canonical stroke positions (\ie, locations before shifting by the framework) and their deformations.
Figure~\ref{fig:method_sketch} shows the overall pipeline to reconstruct dynamic sketches from input video frames.

Before drawing a moving sketch, we load the guided motion field learned in \Sec~\ref{subsec:guiding}.
To initialize curves, we sample points via FPS from the outlier-filtered point cloud at $t=0$ as the first control points.
These points serve as reference markers (\ie, indices) of each curve, and we use their positional encodings as the input of the transformation networks $\mathcal{M}_{\mathrm{R}}$ and $\mathcal{M}_{\mathrm{T}}$ for stroke-wise deformation, along with temporal encodings.
The remaining control points are generated progressively by adding a base radius (\eg $r = 0.02$ for \textit{squat} scene) plus random offsets, maintaining a minimum distance ($\delta = 1.0\times10^{-3}$) between consecutive points.
We initialize $\mathcal{M}_{\mathrm{T}}$ using the weights from our previously trained MLP (described in \Sec~\ref{subsec:guiding}), since both networks handle per-stroke translation.

We model stroke movements as the composition of two components: (1) per-stroke rigid transformations controlling position and orientation and (2) control point adjustments that manage shape changes.
Each component is computed through its corresponding MLP.
The rotation $R_{i}^t\in \mathrm{SO}(3)$ and translation $T_{i}^t\in \mathbb{R}^{3}$ of the $i$-th stroke at the time step $t$ is computed as follows:
\begin{equation}\label{eq:sketch_transform}
    R_{i}^t = \zeta(\mathcal{M}_{\mathrm{R}}(\gamma(P_{i}), \gamma(t))), T_{i}^t = \mathcal{M}_{\mathrm{T}}(\gamma(P_{i}), \gamma(t))
\end{equation}
where $\gamma(\cdot)$ indicates the positional encoder as in \Sec~\ref{subsec:guiding}, and $\zeta(\cdot)$ represents the conversion of angles from quaternion to rotation matrix.
Then, each control point of the stroke is transformed as ${q^{t}}_{i}^{j} = R_{i}^{t}p_{i}^{j}+T_{i}^{t}$ where $s_{i}=\{p_{i}^{j}\}$.
Using these relocated strokes, we compute the changes in control point positions $\{\Delta {p^{t}}_{i}^{j}\}$ as the following equation to deform each curve's control points:
\begin{equation}
    \Delta {p^{t}}_{i}^{j} = \mathcal{M}_{\mathrm{L}}(\gamma({q^{t}}_{i}^{j}), \gamma(t)),
\end{equation}
where $\mathcal{M}_{\mathrm{L}}$ is another network to compute additional changes of each control point.
This optimization process employs a coarse-to-fine approach to efficiently capture global and local features.
Initially, the method operates at low resolution to identify the overall structure and key patterns by learning $\mathcal{M}_{\mathrm{R}}$ and $\mathcal{M}_{\mathrm{T}}$ in the coarse stage.
As the resolution increases, we train $\mathcal{M}_{\mathrm{L}}$ to capture finer details in the fine stage progressively.
The canonical location of control points $\{p_{i}^{j}\}$ are optimized throughout all stages.

The final 3D sketch at the time step $t$ is:
\begin{equation}\label{eq:sketch_fn}
    \mathcal{S}_{\mathrm{3D}}^{t}=\mathcal{S}_{\mathrm{3D}}^{0}+\xi(\Delta \mathcal{S}_{\mathrm{3D}}^{t}),
\end{equation}
where $\mathcal{S}_{\mathrm{3D}}^{0}$ is the canonical state and $\Delta \mathcal{S}_{\mathrm{3D}}^{t}$ represents the accumulated changes throughout the framework.
The function $\xi(\mathbf{x})=\frac{\mathbf{x}}{1+\exp(-a(|\mathbf{x}|-b))}$, with constants $a$ and $b$, suppresses small movements during static periods.
We demonstrate the effect of this correction in \App.

We use the following perceptual loss in the whole process since we need semantic understanding to convey sketches~\cite{vinker2022clipasso}, which is the same as 3Doodle~\cite{choi20243doodle}:
\small
\begin{equation}\label{eq:sketch_frame}
    \mathcal{L}_{\mathrm{frame}}^{\mathrm{s}} = \lambda_{\mathrm{s}}\rho(\mathrm{LPIPS}(\mathcal{I}, \mathcal{S}), \alpha, c) + \mathrm{dist}(\mathrm{CLIP}(\mathcal{I}), \mathrm{CLIP}(\mathcal{S})),
\end{equation}
\normalsize
where $\mathrm{CLIP}(\cdot)$ symbolizes extracted from CLIP~\cite{radford2021learning} image encoder and $\mathrm{dist}(x, y)=1-\frac{x\cdot y}{\|x\|\cdot\|y\|}$ indicates the cosine distance.
We adopt the same parameter values ($\alpha=1$ and $c=0.1$) of the robust function $\rho$ as in \Sec~\ref{subsec:guiding}.

As mentioned in \Sec~\ref{subsec:guiding}, we additionally use regularization terms to enhance quality.
For velocity continuity, we define the following objective function with the overall changes $\Delta \mathcal{S}_{\mathrm{3D}}^{t}$:
\begin{equation}\label{eq:sketch_temporal}
    \mathcal{L}_{\mathrm{temp}}^{\mathrm{s}}=\lambda_{\mathrm{t}}\lVert\frac{\Delta\mathcal{S}_{\mathrm{3D}}^{t}-\Delta\mathcal{S}_{\mathrm{3D}}^{t'}}{t-t'}\rVert_{2},
\end{equation}
where $t'$ is a a neighboring time step of $t$ as same in \Eq~\ref{eq:guiding_temporal}.
The other term for stabilized structure is as follows:
\begin{equation}\label{eq:sketch_reg}  
    \mathcal{L}_{\mathrm{reg}}=
    \begin{cases}
        \lambda_{\mathrm{r}}(\|\smash{^{q}R_{i}}-\smash{^{q}I}\|_{2}+\|T_{i}\|_{2}), & \text{if \textit{coarse} stage}\\
        \lambda_{\mathrm{l}}\|\Delta p_{i}^{j}\|_{2}, & \text{otherwise}
    \end{cases}
\end{equation}
where $R_{i}$ and $T_{i}$ are the per-stroke rotation and translation, and $\Delta p_{i}^{j}$ is the local movement of the control point computed by $\mathcal{M}_{\mathrm{L}}$.
In addition, $\smash{^{q}M}$ denotes the quaternion representation of the matrix $M$, as mentioned in \Sec~\ref{subsec:guiding}.

Finally, we draw the motion as sketches using the following objective function:
\begin{equation}
    \mathcal{L}_{\mathrm{sketch}}=\mathcal{L}_{\mathrm{frame}}^{\mathrm{s}}+\mathcal{L}_{\mathrm{temp}}^{\mathrm{s}}+\mathcal{L}_{\mathrm{reg}}.
\end{equation}
We further discuss the effect of each term in \Sec~\ref{subsec:exp_abl}.

%% file: sec/4_experiment.tex
\section{Experiment}\label{sec:exp}

\input{tables/4_quanti_all}

\paragraph{Implementation Details}
We optimize all networks and sketch parameters using Adam optimizer~\cite{kingma2014adam}.
Moreover, in \Sec~\ref{subsec:guiding}, we learn motion guidance starting from a point cloud with $10k$ points.
We assign values for \Eq~\ref{eq:guiding_total} as $\omega_{\mathrm{f}}=0.1$, $\omega_{\mathrm{t}}=0.05$, and $\omega_{\mathrm{r}}=1.0\times 10^{-4}$.
Similarly, for the equations in \Sec~\ref{subsec:drawing}, we set the following parameters to each term of the objective function: $\lambda_{\mathrm{s}}=0.01$ for \Eq~\ref{eq:sketch_frame}, $\lambda_{\mathrm{t}}=0.01$ for \Eq~\ref{eq:sketch_temporal}, and $\lambda_{\mathrm{r}}=\lambda_{\mathrm{l}}=1.0\times 10^{-3}$ for \Eq~\ref{eq:sketch_reg}.
Moreover, we set $a=100$ and $b=0.05$ for the correction function $\xi(\cdot)$ in \Eq~\ref{eq:sketch_fn}.
Our perceptual distance in~\cref{eq:guiding_frame,eq:sketch_frame} employ VGG16 model for LPIPS loss~\cite{zhang2018perceptual}.
We also use features from a pretrained RN101 model of CLIP encoder~\cite{radford2021learning} for \Eq~\ref{eq:sketch_frame} in sketch synthesis.
Users can control the level of detail by setting the number of strokes.
More implementation details can be found in the Appendix.
\vspace{-8pt}

\paragraph{Datasets}
The inputs are video frames capturing moving objects from a non-stationary camera, along with viewpoints and timesteps.
Since the D-NeRF~\cite{pumarola2021d} dataset lacks camera motion, we rendered a new synthetic dataset with a consistent camera trajectory.
Each scene includes 100 frames with ground-truth camera parameters and images, and all viewpoints are on an upper hemisphere enclosing the target objects.
We also evaluate our model on the real-world scenes from ~\cite{johnson2023unbiased}, ~\cite{liu2024dynamic}, and ~\cite{park2021nerfies}, videos where the camera captures the object while in motion.
\vspace{-8pt}

\paragraph{Baselines}
To the best of our knowledge, we are the first to abstract objects in motion as sketches in 3D space.
Hence, we mainly assess our results by comparing with three baseline methods with varying input/output stroke representations.
Two of them aim to generate 2D sketches.
CLIPasso~\cite{vinker2022clipasso} create an abstract sketch with 2D strokes from a single image, and Zheng \etal~\cite{zheng2024sketch} introduce a framework that generates dynamic sketches from videos.
On the other hand, Suggestive Contours~\cite{decarlo2003suggestive} derives capture geometric contour features from 3D mesh.
Since we have only a 2D video sequence, we first extract meshes using the state-of-the-art dynamic mesh reconstruction~\cite{liu2024dynamic} for this baseline.

Additionally, we compare GS-based~\cite{kerbl20233d} dynamic scene reconstruction works~\cite{wu20244d, yang2023deformable3dgs, huang2023sc,liu2024dynamic} to check how well the output describes the motion.
Note that DG-Mesh~\cite{liu2024dynamic} aims to extract mesh starting from Gaussian splatting techniques, while the others mainly focus on realistic dynamic scene reconstruction in 3D space.

\input{sec/4-1_quantitative}
\input{sec/4-2_qualitative}
\input{sec/4-3_ablation}

%% file: tables/4_quanti_all.tex
\begin{table*}[ht]\setlength{\tabcolsep}{3pt}
\vspace{-4pt}
\setlength{\abovecaptionskip}{7pt}
\footnotesize
\centering
\begin{tabular}{c c}
\hspace{-5pt}\raisebox{-2.5pt}{
\begin{tabular}{lcccc}
    \toprule
    \multirow{2}{*}{Method} & \multicolumn{2}{c}{Strcutural alignment $\left(\uparrow\right)$} & \multicolumn{2}{c}{Motion prompt similarity $\left(\uparrow\right)$} \\
    & Novel views & Fixed views & Novel views & Fixed views \\
    \midrule 
    CLIPasso & $0.760 \pm 0.107$ & $0.740 \pm 0.127 $ & $0.659 \pm 0.007$ & $0.664 \pm 0.011$ \\
    Sketch Video Syn. & $0.663 \pm 0.115$ & $0.657 \pm 0.135$ & $0.654 \pm 0.011$ & $0.658 \pm 0.011$ \\
    Sugg. Contours & $0.784 \pm 0.102$ & $0.750 \pm 0.119$ & $0.661 \pm 0.013$ & $0.656 \pm 0.016$ \\
    \midrule
    Liv3Stroke (Ours) & $0.693 \pm 0.096$ & $0.683 \pm 0.108$ & $0.656 \pm 0.006$ & $0.656 \pm 0.008$ \\
    \bottomrule \\[5pt]
    \multicolumn{5}{c}{(a) Quantitative results of dynamic 3D sketches.}
\end{tabular}
}

&

\hspace{2.5pt}\begin{tabular}{lcc}
    \toprule
    Method & \makecell{Per-frame \\ Chamfer $\left(\downarrow\right)$} & \makecell{Motion velocity \\ distance ($\times 10^{-3}) \left(\downarrow\right)$} \\
    \midrule 
    4DGS & $0.205 \pm 0.046$ & $4.60  \pm 3.00$  \\
    Deformable 3DGS & $0.302 \pm 0.170$  &  $5.05 \pm 4.06$  \\
    SC-GS & $0.294 \pm 0.055$  & $ 4.21 \pm 2.75 $ \\
    DG-Mesh & $0.277 \pm 0.059$  & $4.10 \pm 2.70$  \\
    \midrule
    Liv3Stroke (Ours) & $0.252 \pm 0.049$  & $4.16 \pm 2.34$\\
    \bottomrule \\
    \multicolumn{3}{c}{(b) Quantitative results of 3D guidance accuracy.}
\end{tabular}

\end{tabular}
\caption{
    Overall quantitative results.
    \textbf{(a)} Quantitative metrics on sketch video frames.
    We evaluate structural alignment score with edge images of referenced frames and CLIP feature-based motion prompt similarity.
    \textbf{(b)} Quantitative results of 3D guidance accuracy.
    Our approach can capture meaningful 3D motion information as the point cloud sequence, even though we do not pursue realistic scenes.
}
\label{tab:4_all_quant}
\vspace{-5pt}

\end{table*}

%% file: sec/4-1_quantitative.tex
\subsection{Quantitative Results}\label{subsec:exp_quanti}

We provide quantitative results of sketch videos in \tab~\ref{tab:4_all_quant} (a).
In this table, we evaluate how well the sketch preserves the overall structure and the desired movement.
We evaluate both aspects from novel camera trajectories and fixed viewpoints.
For structure preservation, we employ MS-SSIM~\cite{wang2003multiscale} metrics between generated sketch frames and their corresponding reference edge views, which are extracted from PiDiNet~\cite{su2021pixel}.
For motion expression quality, we compute CLIP~\cite{radford2021learning} feature similarities between each frame and the text prompt structured as ``A sketch of \{\textit{movement}\}'', and normalize values to $[0, 1]$.
Note that we use ViT-B/32~\cite{alexey2020image} models in this evaluation, which remain independent from our optimization pipeline to ensure unbiased evaluation.

We observe that our approach has the smallest deviations in both scenarios, showing that Liv3Stroke can consistently capture the structure throughout the whole input sequence.
Sketch Video Synthesis achieves the lowest scores in this validation since it mainly focuses on depicting detailed features rather than capturing the whole movement. 
While CLIPasso performs well in this validation, its higher deviations show that it lacks consistency in structure throughout the sequence.
Similarly, Suggestive Contours achieve the highest scores, as it tend to generate outline contours as illustrated in \fig~\ref{fig:exp_quali_all}.
We further discuss about this in \Sec~\ref{subsec:exp_qual}.

In motion prompt similarity scores, Liv3Stroke achieves consistent scores with the smallest deviation in both novel camera trajectory and fixed viewpoint scenarios compared to other approaches.
This demonstrates that our approach maintains stable performance in motion expression while being robust to varying camera viewpoints.
Furthermore, Sketch Video Synthesis achieves the lowest score in the camera-moving scenario compared to the second-highest score when the camera is stationary.
This is because they are dependent on layered neural atlas~\cite{neuralatlas}, which cannot represent big movements.

We also evaluate motion guidance performance described in \Sec~\ref{subsec:guiding}. Table~\ref{tab:4_all_quant} (b) presents motion accuracy comparisons between GS-based dynamic approaches~\cite{wu20244d, yang2023deformable3dgs, huang2023sc, liu2024dynamic} and our method, measured on our synthetic dataset.
Using point clouds extracted from mesh at each time step as ground truth, we assess the performance through two metrics: (1) per-frame structural accuracy measured by Chamfer distance between point clouds and (2) motion velocity accuracy calculated by L2 loss between ground truth and predicted point position changes at each time step.
Although 4DGS achieves better per-frame structural accuracy scores than ours, their overall structure deteriorates as shown in \fig~\ref{fig:exp_quali_guidance}.
We discuss this phenomenon further in \Sec~\ref{subsec:exp_qual}.

Meanwhile, Liv3Stroke performs similarly to DG-Mesh but with fundamentally different objectives.
While DG-Mesh focuses on precise mesh reconstruction, Liv3Stroke aims for efficient motion capture by simply repositioning point clouds to obtain approximate 3D motion information. 
This distinction in goals highlights the efficiency of our approach since we achieve similar quantitative results despite using a simpler approach focused on motion rather than detailed geometry.

%% file: sec/4-2_qualitative.tex
\subsection{Qualitative Results}\label{subsec:exp_qual}

\begin{figure*}
    \centering
    \includegraphics[trim={0mm 0mm 2mm 0mm}, clip, width=0.99\linewidth]{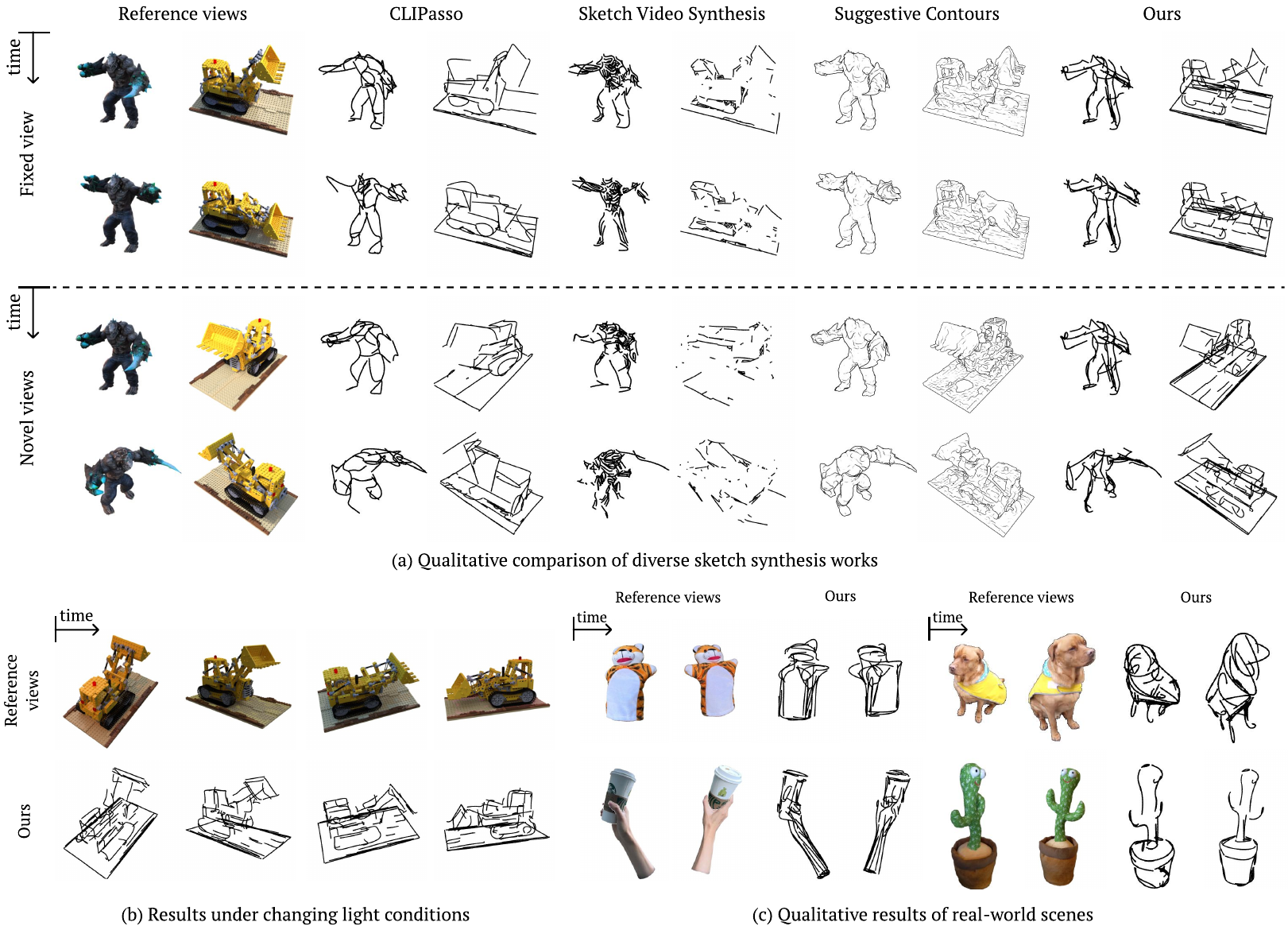}
    \vspace{-2pt}
    \caption{Qualitative results of sketches.
    \textbf{(a)} Comparison with baselines.
    Our approach can generate sketches with view-consistent motions from RGB frames simply by locating and deforming each stroke.
    Note that Suggestive Contours~\cite{decarlo2003suggestive} requires the input mesh, hence cannot capture motions directly from videos.
    \textbf{(b)} Liv3Stroke also can represent 3D movements even when RGB values change due to the surrounding environment.
    \textbf{(c)} Results of real-world scenes.
    Our approach successfully captures movements in real-world scenarios.
    }
    \label{fig:exp_quali_all}
    \vspace{-5pt}
\end{figure*}

\begin{figure}[t]
    \centering
    \includegraphics[trim={2mm 0mm 3mm 0mm}, clip, width=0.93\linewidth]{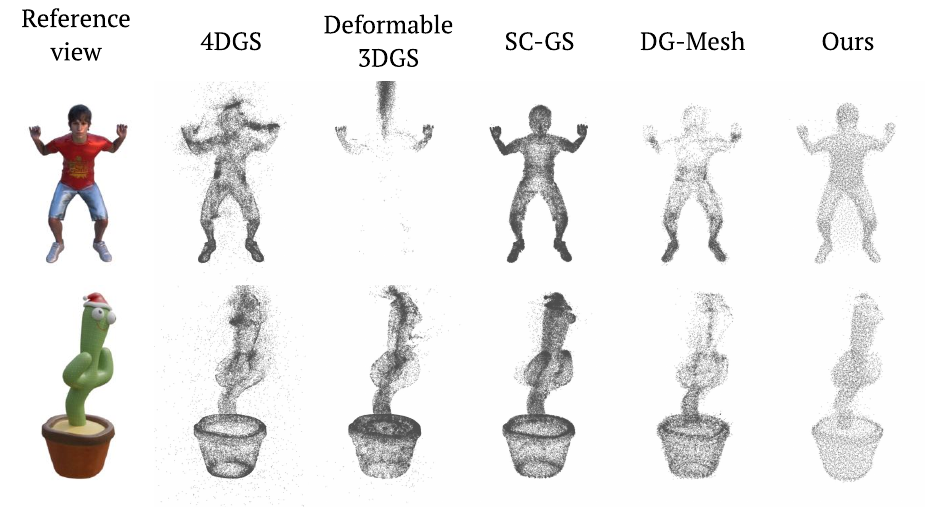}
    \vspace{-5pt}
    \caption{Qualitative results on guidance.
    Our approach captures clearer structure compared to photorealistic reconstruction methods despite not primarily aiming for accurate scene reproduction.
    }
    \label{fig:exp_quali_guidance}
    \vspace{-15pt}
\end{figure}

We provide the qualitative results of our sketches in \cref{fig:teaser,fig:exp_quali_all}.
In \fig~\ref{fig:exp_quali_all} (a), we compare our method against existing baselines under both fixed viewpoints and novel camera trajectories.

From these scenarios, our method exhibits several key advantages.
CLIPasso demonstrates inconsistent structural details and unstable stroke placements across all scenarios.
While Sketch Video Synthesis achieves better temporal coherence from fixed viewpoints, it struggles to maintain structural integrity during camera motion.
These limitations stem from their primary focus on 2D synthesis, resulting in an insufficient understanding of 3D motions.
Furthermore, it struggles with the \textit{lego} scene in \fig~\ref{fig:exp_quali_all} (a), as it tends to produce overly detailed representations instead of focusing on essential features.
Although Suggestive Contours generates more structured results, it requires meshes for drawing, making it heavily rely on the quality of input meshes.
Additionally, the results tend toward detailed depiction rather than abstraction.
In contrast, our approach successfully represents movements in both fixed and moving camera scenarios, while maintaining consistent structural integrity throughout the motion sequence.
The advantage of our method becomes particularly evident in novel camera trajectories, where it robustly preserves 3D structural features across different viewpoints.

Figure~\ref{fig:exp_quali_all} (b) shows our sketch representation's robustness to lighting conditions, a significant external factor that affects frame RGB values.
Liv3Stroke maintains consistent sketch video generation while preserving core structure and motion, even with varying lighting colors throughout the sequence.

In addition, our approach can also effectively represent real-world scenarios as in \fig~\ref{fig:exp_quali_all} (c).
It can capture various objects and their movements.
Despite the complexity of real-world scenes, it preserves key structural characteristics of each object.
We highly recommend finding the supplement to see more results.

Meanwhile, \fig~\ref{fig:exp_quali_guidance} compares our point cloud reconstruction with other methods.
We observe that 4DGS exhibits noisy point distributions and Deformable 3DGS struggles with overall structure preservation, particularly visible in the elongated artifacts.
While DG-Mesh demonstrates higher performance in 3D shape reconstruction due to its mesh-focused approach, it shows concentrated point distribution in certain areas.
Compared to these methods, our method successfully maintains structural integrity during object motion with uniform point distribution despite optimizing for image intensity rather than realistic reconstruction objectives.

%% file: sec/4-3_ablation.tex
\subsection{Ablation study}\label{subsec:exp_abl}

\begin{figure}
    \centering
    \includegraphics[trim={0mm 0mm 0mm 0mm}, clip, width=1.0\linewidth]{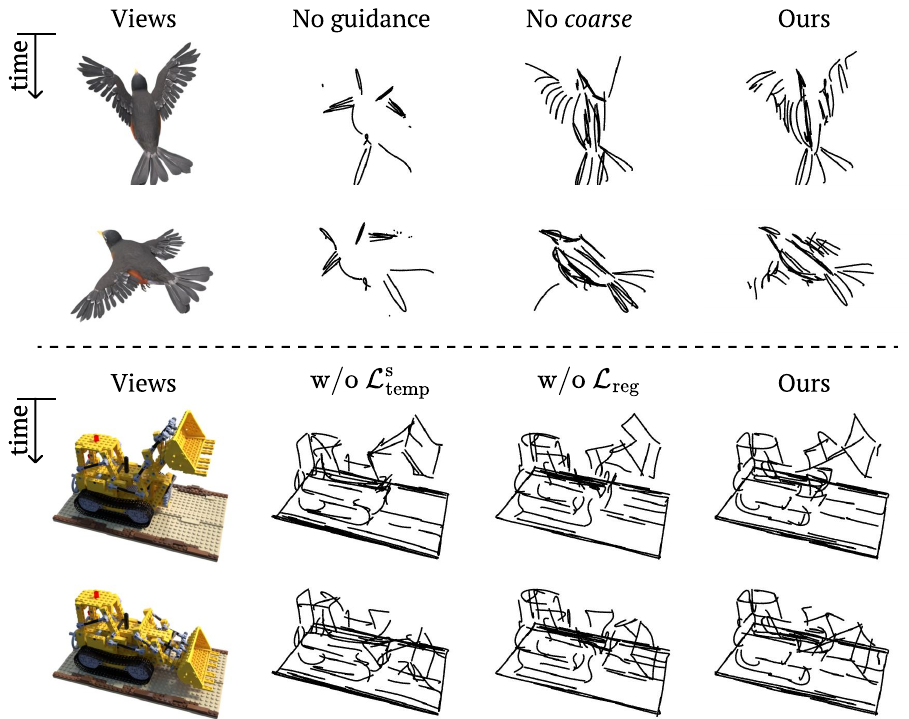}
    \vspace{-2pt}
    \caption{Ablation study on design choices in sketch reconstruction.
    }
    \label{fig:exp_abl_sketch}
    \vspace{-5pt}
\end{figure}

We further validate the details of our framework through an ablation study.
Figure~\ref{fig:exp_abl_sketch} shows the qualitative results of design choices in generating sketches.

As observed, directly optimizing strokes without guidance leads to disconnected line segments that barely capture the motion.
Without coarse guidance, while the overall shape is preserved, the output lacks coherent structure and key features, such as wings of the bird.
Our full framework generates the most balanced results, maintaining global postures and fine details.
Similarly, it is difficult to capture the key motion when learning without $\mathcal{L}_{\mathrm{temp}}^{\mathrm{s}}$.
As shown in \fig~\ref{fig:exp_abl_sketch}, we observe unstable stroke variations in the generated sketches even when there are no changes between input frames.
Without $\mathcal{L}_{\mathrm{reg}}$, although motion dynamics are preserved, the model struggles to capture the essential structural characteristics of the object.
Compared to these results, our complete model reconstructs sketches with the best structural representation and expresses stable movements, demonstrating how each component contributes to the final sketch quality.

\input{tables/4_abl_motion}

We also conduct an ablation study to analyze the effectiveness of each component in our motion guidance generation framework, as described in \Sec~\ref{subsec:guiding}.
Table~\ref{tab:abl_motion_acc} shows the quantitative results measuring per-frame structure and motion velocity accuracy.
Our full framework achieves the best performance among the conditions related to terms in the objective function.
Without $\mathcal{L}_{\mathrm{temp}}^{\mathrm{g}}$, the model shows degraded performance, especially in motion velocity distance per time step, indicating the importance of temporal coherence in extracting motion guidance.
When removing $\mathcal{L}_{\mathrm{rigid}}$, we observe similar performance in structural similarity but decreased motion velocity accuracy.
We can interpret this that $\mathcal{L}_{\mathrm{rigid}}$ helps capture consistent motion patterns.
Using L2 loss instead of our proposed $\mathcal{L}_{\mathrm{rigid}}$ function shows a slightly high motion velocity distance, demonstrating the effectiveness of our rigid loss formulation.
Furthermore, our approach ensures the accuracy of both 3D structure and motion when using RGB images instead of masks or grayscale frames.
We provide more comparisons in \App.

%% file: tables/4_abl_motion.tex
\begin{table}[t]
    \centering
    \resizebox{0.8\linewidth}{!}{
    \begin{tabular}{@{}lcc@{}}
        \toprule
          & \makecell{Per-frame \\ Chamfer $\left(\downarrow\right)$} & \makecell{Motion velocity \\ distance ($\times 10^{-3}) \left(\downarrow\right)$} \\
        \midrule 
        w/o $\mathcal{L}_{\mathrm{temp}}^{\mathrm{g}}$ & $0.258 \pm 0.043$ & $6.81 \pm 5.41$  \\
        w/o $\mathcal{L}_{\mathrm{rigid}}$ & $0.253 \pm 0.048$ & $5.45 \pm 3.74$  \\
        L2 in $\mathcal{L}_{\mathrm{rigid}}$ & $0.253 \pm 0.045$  & $4.61 \pm 2.94$ \\
        \midrule
        Ours (mask) & $0.257 \pm 0.049$ & $4.12 \pm 2.84$ \\
        Ours (grayscale) & $0.249 \pm 0.039$ & $4.41 \pm 2.46$ \\
        \midrule
        Liv3Stroke (Ours) & $0.252 \pm 0.049$  & $4.16 \pm 2.34$ \\
        \bottomrule
    \end{tabular}
    }
    \caption{Ablation study on generating motion guidance.}
    \vspace{-10pt}
    \label{tab:abl_motion_acc}
\end{table}

%% file: sec/5_conclusion.tex
\section{Conclusion}
In this work, we introduce Liv3Stroke, a novel approach that bridges the gap between video analysis and motion abstraction by representing 3D object movements through dynamic stroke manipulation.
Our method demonstrates that diverse movements can be effectively represented with sparse strokes by relocating and deforming them.

Looking forward, our work opens new possibilities for understanding scene flow dynamics and estimating large-scale motion fields, potentially advancing computer vision and motion analysis domains.
In addition, building on our approach's representation of motion through 3D strokes, stroke-based physical control could be achieved by discretizing the dense field and designing functions that map physical properties between strokes and the field.

\paragraph{Limitations}
Since we only consider view-independent strokes, our approach cannot render view-dependent representations such as a contour of the rounded object.
We expect to overcome these limitations by adopting view-dependent strokes with superquadrics as proposed in 3Doodle~\cite{choi20243doodle}.

%% file: sec/9_ack.tex
\subsection*{Acknowledgments}
This work was supported by IITP grant (RS-2021-II211343: AI Graduate School Program at Seoul National University) (5\%) and No.2021-0-02068: Artificial Intelligence Innovation Hub (45\%)) and NRF grant (No. 2023R1A1C200781211 (55\%)) funded by the Korea government (MSIT).

%% file: sec/X_suppl.tex

\setcounter{section}{0}
\renewcommand{\thesection}{\Alph{section}}
\setcounter{figure}{0}
\renewcommand{\thefigure}{\Alph{figure}}
\setcounter{table}{0}
\renewcommand{\thetable}{\Alph{table}}

\maketitlesupplementary

\section{Methodological Details}


\subsection{Point Cloud Rasterization}\label{sup:rasterize}

To rasterize the point cloud to an image plane in \Sec~\ref{subsec:guiding}, we first project the 3D points $P_{3D}=\{P_{i}\}$ onto the 2D space $P_{2D}=\{\tilde{P}_{i}\}$.
For each pair of $i$-th normalized grid point and $j$-th point of the normalized point cloud $\hat{P}_{2D}$ to image dimensions, we use the Gaussian function to compute a rendered intensity $J_{ij}$:
\begin{equation}
    J_{ij}=\exp{(-\frac{D_{ij}^2}{2\sigma_{j}^{2}})},
\end{equation}
where $D_{ij}$ is the Euclidean distance between two points and $\sigma_{j}$ indicates the point size factor that controls the contribution area of each point.

We dynamically adjust $\sigma_{j}$ based on the depth of the point $\{d_{j}\}$ to account for perspective projection effects.
Points farther from the camera are rendered with smaller sizes, following standard 3D rendering principles. Using the normalized depth $\{\hat{d}_{j}\}=\{\frac{d_{j}-d_{min}}{d_{min}-d_{max}}\}$, each point size $\sigma_{j}$ is computed as:
\begin{equation}
    \sigma_{j}=\frac{\mu}{0.5 \mu \min{(W, H)}}\times \hat{d}_{j} \times \beta,
\end{equation}
where $\mu$ and $\beta$ denote the scaling and deblurring factor, and $W$, $H$ are the image width and height.
We set $\mu=10$ and $\beta=0.5$.

We aggregate the Gaussian contributions from all point cloud points to each grid point to generate the final rendered image.
The intensity value for each pixel is computed by summing these contributions.
We then normalize the intensities by dividing by the maximum value, ensuring the final image $\mathcal{J} \in \mathbb{R}^{H\times W}$ values fall within an appropriate range for visualization or processing.
This process can be expressed as the following equation:
\begin{equation}
    \mathcal{J}_{i}=\frac{\sum_{j=1}^{M}R_{ij}}{\max(\sum_{j=1}^{M}R_{ij})}.
\end{equation}
The results can be shown in \cref{fig:supp_implement_res,fig:supp_all_1}.
Note that each guidance view in \cref{fig:supp_implement_res} is rasterized into a $100\times100$ resolution image, which represents the actual resolution used for generating motion guidance in synthetic scenes.

\subsection{Effect of the Suppression Function $\xi(\cdot)$}

\begin{figure}
    \centering
    \includegraphics[trim={5mm 3mm 0mm 0mm}, clip, width=0.85\linewidth]{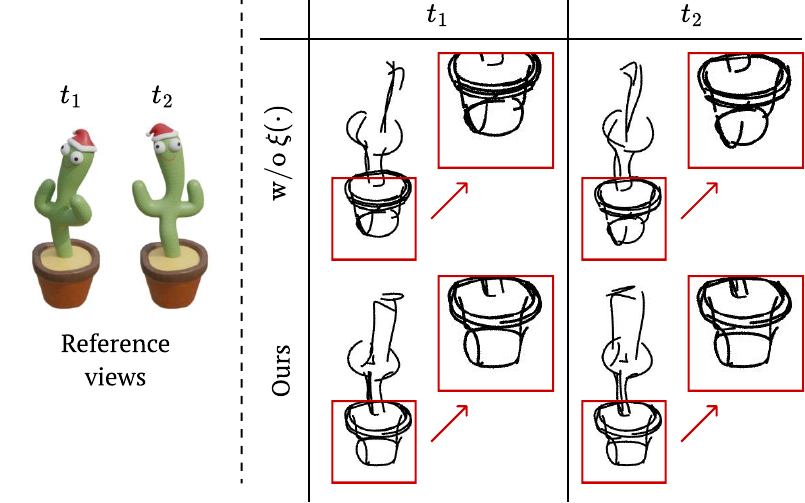}
    \caption{Effectiveness of the function for suppression $\xi(\cdot)$.
    Without motion suppression, we observe noisy stroke movement at different time steps ($t_1$ and $t_2$) even if there are no motions.
    }
    \label{fig:supp_abl_xi}
    \vspace{-5pt}
\end{figure}

As described in \Sec~\ref{subsec:drawing}, we adjust the suppression function $\xi(\cdot)$ to prevent unintended stroke movements in sketch synthesis.
Figure~\ref{fig:supp_abl_xi} shows the effect of this function.
We observe that the model struggles to suppress undesired stroke movements even when no motion occurs.
This demonstrates that our full approach achieves higher performance in extracting core motions.

\section{Implementation Details}\label{sup:implementation}

\subsection{Network Architecture}

\begin{figure}
    \centering
    \includegraphics[width=0.85\linewidth]{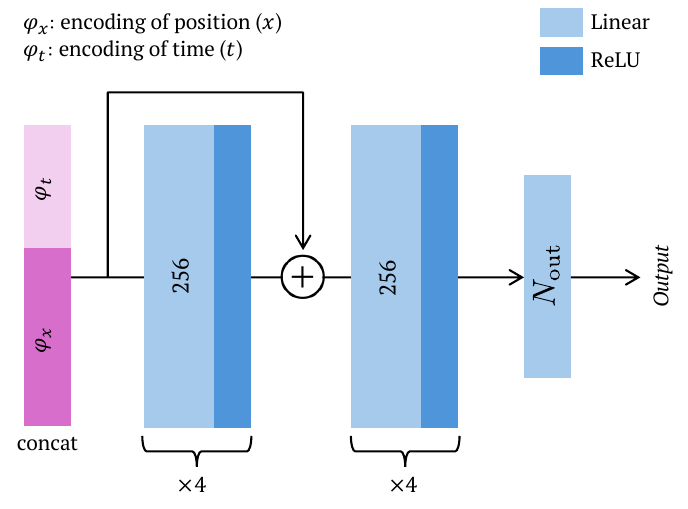}
    \vspace{-7pt}
    \caption{Network architecture. All networks in the framework share the same architecture. $N_{\mathrm{out}}$ indicates the dimension of the output, which is $4$ in $\mathcal{M}_{\mathrm{R}}$, and $3$ in the others.
    }
    \label{fig:supp_network}
    \vspace{-10pt}
\end{figure}

Our network architecture, illustrated in \fig~\ref{fig:supp_network} follows a consistent MLP structure across all components in \Sec~\ref{subsec:guiding} and ~\ref{subsec:drawing}, adopting a similar design to that proposed by ~\cite{pumarola2021d}.
Input is the concatenation of positional encoding of time $\varphi_{t}$ and positions $\varphi_{x}$, and each linear layer, except for the final layer, outputs a 256-dimensional feature vector.
The network $\mathcal{M}_{\mathrm{R}}$ yields outputs in $\mathbb{R}^{N\times4}$, while other networks produce output vectors in $\mathbb{R}^{N\times3}$.
$\mathcal{M}_{\mathrm{R}}$ outputs quaternions for each stroke's rotation, which are subsequently converted to rotation matrices for stroke deformation.

\subsection{Optimization Details}

\begin{figure}
    \centering
    \includegraphics[trim={0mm 0mm 0mm 0mm}, clip, width=1.0\linewidth]{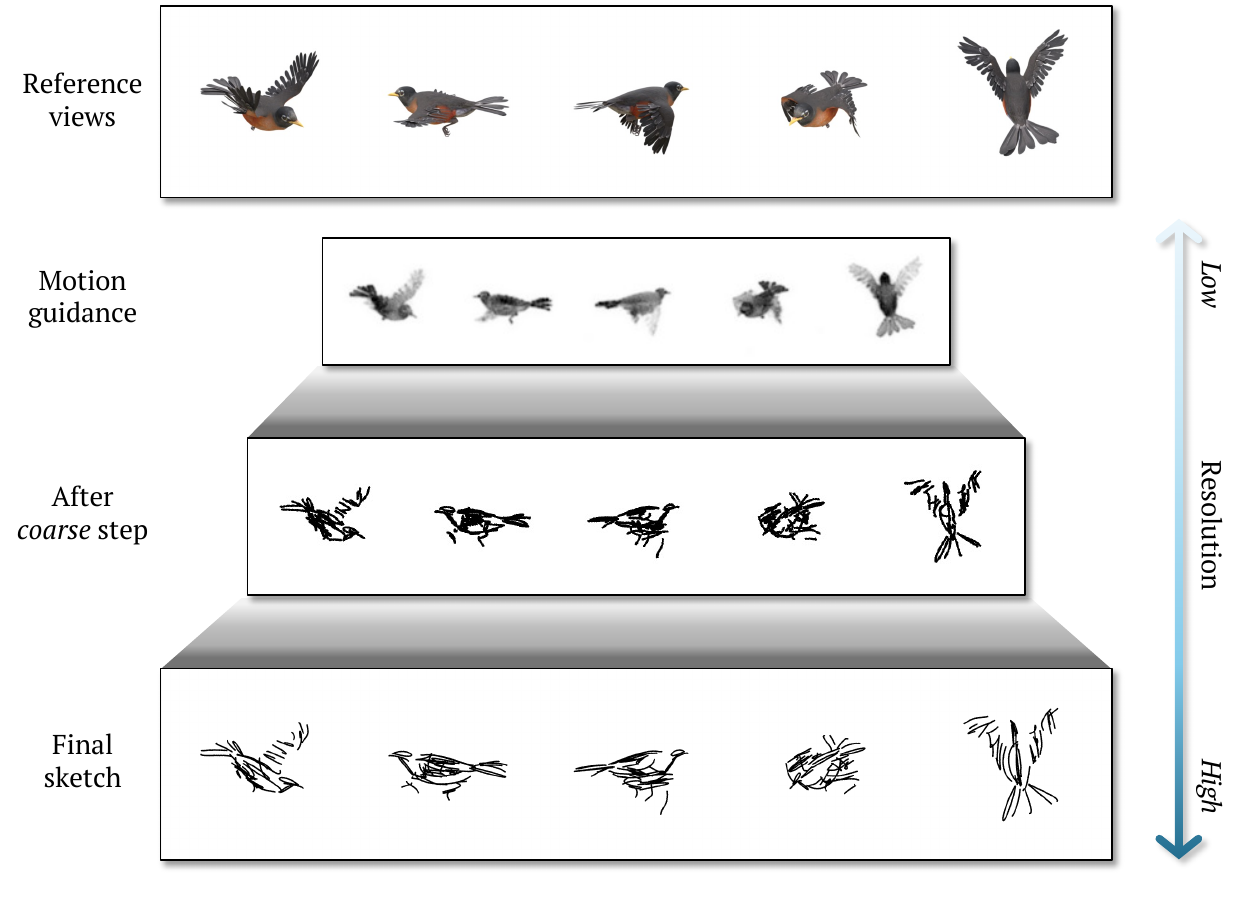}
    \caption{Different resolution through processing stages.
    Our method gradually increases resolution according to stages to compute the location and deformation of strokes.
    }
    \vspace{-5pt}
    \label{fig:supp_implement_res}
\end{figure}

The learning parameters differ between reconstructing synthetic datasets and real-world scenes.
For synthetic scenes, we set the frequency value $L=10$ for both temporal and spatial positional encoding.
For real scenes presented in \fig~\textcolor{cvprblue}{9}, we use $L=8$ for temporal and $L=10$ for spatial encoding.
Learning rate values are also slightly different in synthetic and real scenes.
For the former, in drawing process as described in \Sec~\ref{subsec:drawing}, we apply a learning rate of $5.0\times10^{-4}$ to $\mathcal{M}_{\mathrm{T}}$ and $\mathcal{M}_{\mathrm{R}}$, and $1.0\times10^{-3}$ to all other parameters.
For the latter, during sketch reconstruction, we apply a learning rate of $5.0\times10^{-4}$ to the canonical stroke positions, $1.0\times10^{-4}$ to $\mathcal{M}_{\mathrm{T}}$ and $\mathcal{M}_{\mathrm{R}}$, and $2.5\times10^{-4}$ to $\mathcal{M}_{\mathrm{L}}$.
We set the learning rates $lr_{\mathrm{pcd}}=1.0\times10^{-3}$ and $lr_{\mathrm{mlp}}=5.0\times10^{-4}$ to optimize the canonical point cloud (\ie, the point cloud before network-based shifting) and the motion guidance function detailed in \Sec~\ref{subsec:guiding} for all scenes.
In addition, during learning motion guidance, to embed core motion information into the network, we initialize a canonical point cloud at $t=0$ and reset the network parameters in the middle of the process.

Meanwhile, our framework is structured to gradually increase resolution as the optimization process progresses.
As shown in \fig~\ref{fig:supp_implement_res}, We initially obtain motion guidance at quarter resolution of the target image size.
Then, in the coarse stage of sketch synthesis, we render strokes into a $50\%$ of the full resolution.
We finally get moving sketches by optimizing the full resolution of the target frame size.
For instance, for synthetic scenes, we first learn guidance at $100\times100$ resolution and then optimize the per-stroke transformation using $200\times200$ frames. 
The final output produces sketch frames at $400\times400$ resolution.

\section{Additional Results}

\begin{figure*}
    \centering
    \includegraphics[trim={0mm 0mm 0mm 0mm}, clip, width=0.95\linewidth]{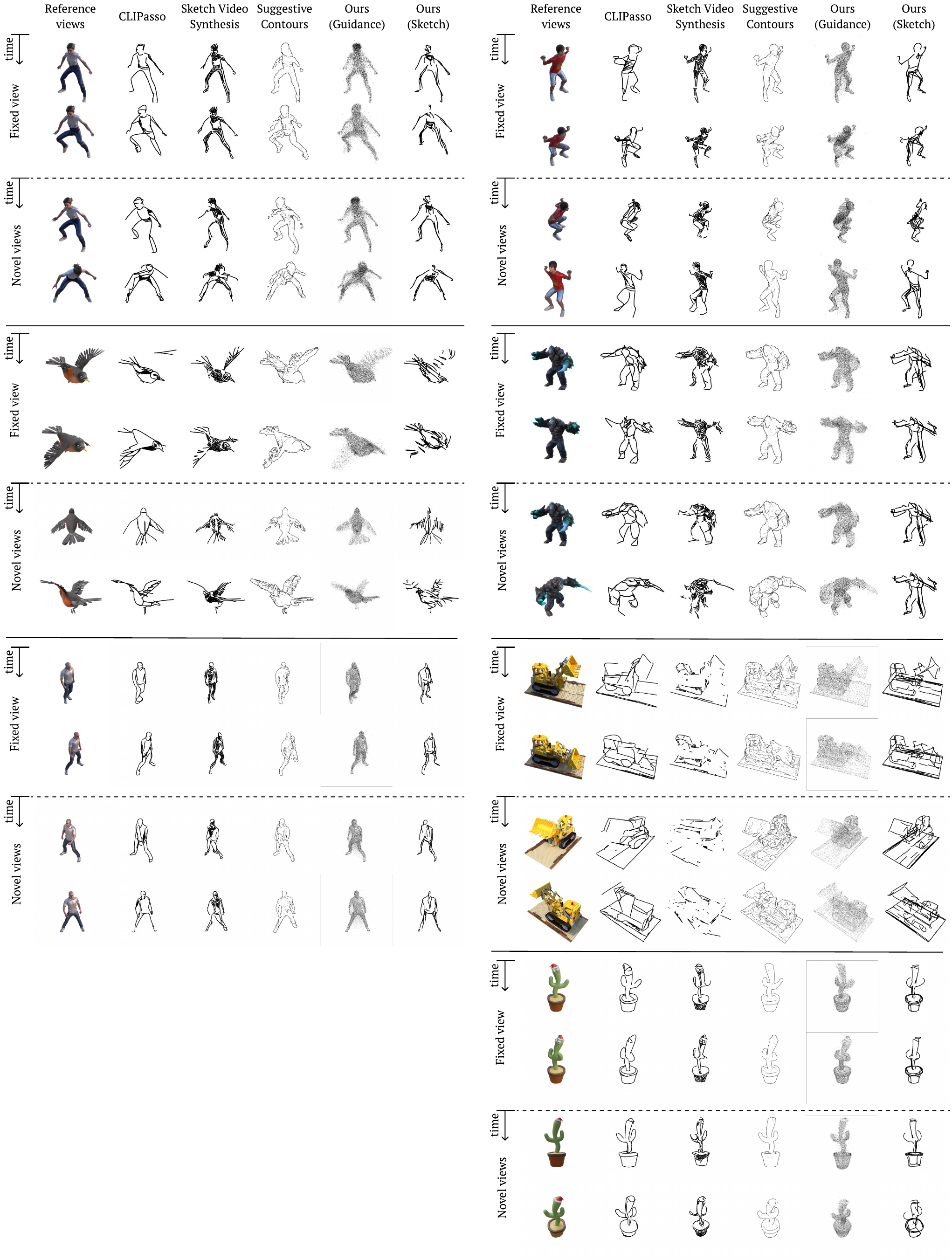}
    \caption{Results of synthetic scenes.
    Our approach can represent diverse motions by using view-consistent deformable 3D strokes.
    }
    \label{fig:supp_all_1}
\end{figure*}

We provide results of all rendered synthetic scenes in \fig~\ref{fig:supp_all_1}.
Compared to other existing works, our framework can represent diverse movements and key features of the view-consistent structure directly from RGB video frames.
We visualize guidance views at the full target image resolution for better clarity.
We highly recommend finding videos in the supplementary material to see the whole movement of each scene.

\subsection{Quantitative Results of the Motion Guidance}

\input{tables/X_quanti_motion}

We present additional quantitative results of the motion guidance that we obtained from \Sec~\ref{subsec:guiding} and filtered results of GS-based dynamic reconstruction works~\cite{wu20244d, yang2023deformable3dgs, huang2023sc} according to the opacity value with a threshold of $\alpha=0.5$.
Table~\ref{tab:supp_guidance} and \tab~\ref{tab:4_all_quant} (b) of the main paper shows our method's capability to capture meaningful 3D motion information, although it does not pursue realistic reconstruction.

\subsection{Results of Different Number of Strokes}

\begin{figure}
    \centering
    \includegraphics[trim={0mm 0mm 0mm 0mm}, clip, width=0.97\linewidth]{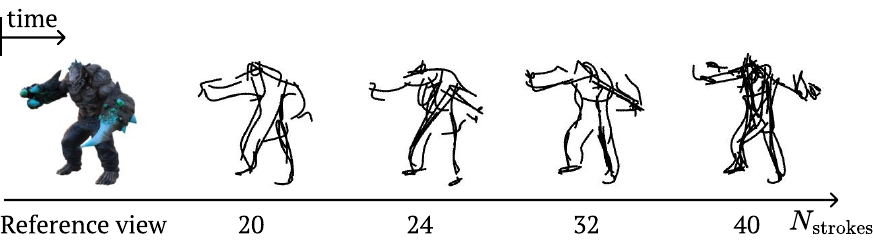}
    \caption{The effects of using different numbers of strokes.
    When reconstructing a sketch video, we allow users to set $N_{\mathrm{strokes}}$.
    More strokes produce detailed sketches, while fewer strokes yield abstract ones.
    }
    \label{fig:supp_abl_n_strokes}
    \vspace{-5pt}
\end{figure}

We provide results to show the effect of the number of strokes as in ~\cref{fig:supp_abl_n_strokes}.
Like \cite{choi20243doodle} and \cite{vinker2022clipasso}, we can control abstraction levels of sketches by adjusting the number of curves.
With a higher number of strokes, we can capture more detailed features, while fewer strokes result in more abstract representations.

\subsection{Limited Multi-View Information}

\begin{figure*}
    \centering
    \includegraphics[trim={0mm 3mm 0mm 0mm}, clip, width=1.0\linewidth]{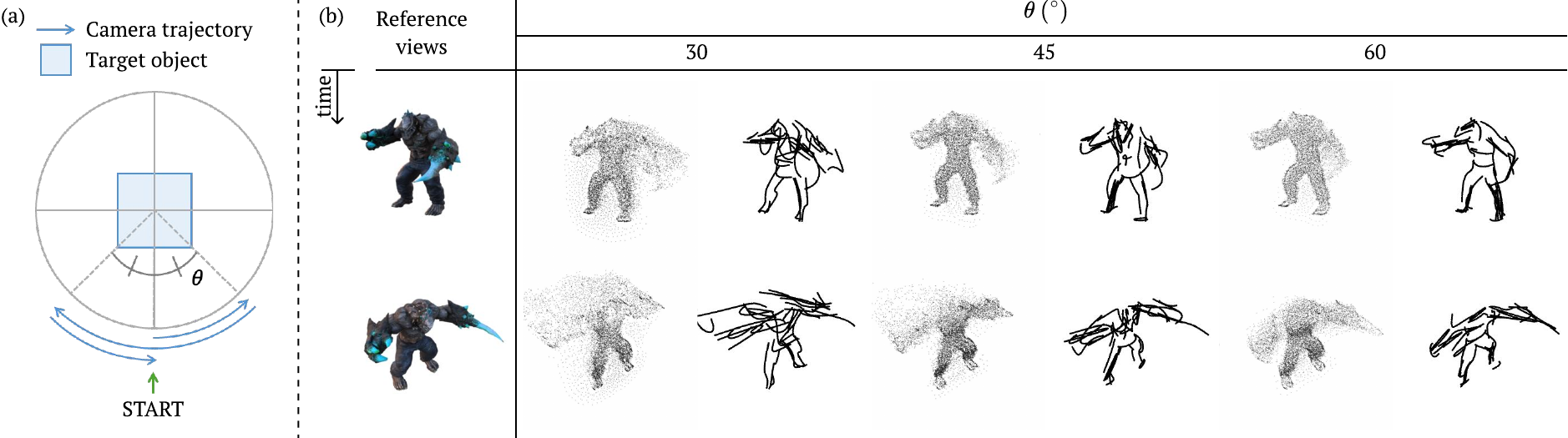}
    \caption{Results of limited multi-view information.
    \textbf{(a)} Experimental setup for data collection.
    We followed a circular path around the object to capture frames, beginning from the frontal view (marked with a green arrow).
    \textbf{(b)} Results of the motion guidance and sketch under different angles.
    Our approach can achieve 3D motion sketch representation with limited yet sufficient viewpoint information ($\theta \geq 45$), even when the motion guidance exhibits noise, such as at $\theta = 45$.
    }
    \label{fig:supp_limited}
\end{figure*}

We provide results under limited multi-view information in varied conditions.
From a frontal view, we captured frames along a circular trajectory around the object, collecting 100 frames over an angle $\theta(^\circ)$ while maintaining a constant distance from the center.
The detailed experimental setup is visualized in ~\cref{fig:supp_limited} (a).

Our approach renders 3D sketches of the object in motion with sparse yet adequate viewpoint information, as illustrated in ~\cref{fig:supp_limited} (b). Additionally, we find that our method can roughly capture the 3D key structure of the object even when the motion guidance exhibits noise, such as at $\theta = 45$.

\subsection{User study}

\input{tables/X_user_study}

We provide a questionnaire to evaluate the perceptual implication of generated sketches.
Participants rated the sketches on a five-point scale (1-5), evaluating them from both novel camera viewpoints and the fixed perspective.
The rating criteria were: (1) how effectively the sketch captures the motion and (2) how well it conveys the 3D structure of the target object.

Table~\ref{tab:supp_user_study} summarizes the answers of 44 participants.
Overall, Suggestive Contours~\cite{decarlo2003suggestive} achieves the highest scores across all metrics, which can be attributed to its direct contour extraction from 3D meshes, as illustrated in \fig~\textcolor{cvprblue}{6}.
Unlike other methods that rely on image-based processing, this approach results in higher evaluation scores.
For novel views, LiveStroke performs comparably with CLIPasso~\cite{vinker2022clipasso}. 
While Sketch Video Synthesis~\cite{zheng2024sketch} has a higher score in the fixed views, it struggles to effectively capture 3D geometric features and motion characteristics when evaluated from moving camera perspectives.
LiveStroke exhibits only minimal performance decrease when transitioning from the novel perspectives to the fixed view, demonstrating consistent performance regardless of the viewing perspective.
This stability distinguishes our approach from others, which shows significant performance variations between different viewpoints.

%% file: tables/X_quanti_motion.tex
\begin{table}
    \centering
    \resizebox{0.82\linewidth}{!}{
    \begin{tabular}{@{}lcc@{}}
        \toprule
           & \makecell{Per-frame \\ Chamfer $\left(\downarrow\right)$} & \makecell{Motion velocity \\ distance ($\times 10^{-3}) \left(\downarrow\right)$} \\
        \midrule 
        4DGS\textsuperscript{\textdagger} & $0.286 \pm 0.057$ & $4.24 \pm 2.71$  \\
        Deformable 3DGS\textsuperscript{\textdagger}  & $0.269 \pm 0.071$ & $4.01 \pm 2.67$  \\
        SC-GS\textsuperscript{\textdagger} & $0.289 \pm 0.053$  & $3.99 \pm 2.65$ \\
        \midrule
        Liv3Stroke (Ours) & $0.252 \pm 0.049$  & $4.16 \pm 2.34$\\
        \bottomrule
    \end{tabular}
    }
    \caption{Quantitative results of 3D motion guidance accuracy of \textsuperscript{\textdagger}GS-based works with filtering based on the opacity value.
    }
    \vspace{-5pt}
    \label{tab:supp_guidance}
\end{table}

%% file: tables/X_user_study.tex
\begin{table}
    \centering
    \resizebox{0.9\linewidth}{!}{
    \begin{tabular}{@{}lcccc@{}}
        \toprule
        \multirow{2}{*}{Method} & \multicolumn{2}{c}{Novel views} & \multicolumn{2}{c}{Fixed views} \\
        & Motion & Structure & Motion & Structure \\
        \midrule 
        CLIPasso & $3.32$ & $3.08$ & $2.87$ & $2.70$ \\
        Sketch Video Syn. & $2.66$ & $2.64$ & $3.93$ & $3.77$ \\
        Sugg. Contours & $4.26$ & $4.29$ & $3.82$ & $3.90$ \\
        \midrule
        Liv3Stroke (Ours) & $3.32$ & $3.08$ & $3.21$ & $2.94$ \\
        \bottomrule \\
    \end{tabular}
    }
    \vspace{-1.0em}
    \caption{User study results. Note that ``Motion'' denotes the evaluation of how well the result describe the desired movement, and ``Structure'' is the score of how well it contains key features of the 3D structure.}
    \vspace{-10pt}
    \label{tab:supp_user_study}
\end{table}